\documentclass{article}
\usepackage{spconf,amsmath,epsfig}
\usepackage{amsmath,amsthm,amssymb,amsfonts}
\usepackage{array, multirow}
\usepackage{multirow}
\usepackage{graphicx}
\usepackage{color}
\usepackage{url}
\usepackage{hyperref}
\usepackage{amsmath}
\DeclareMathOperator*{\argmax}{arg\,max}

\let\OLDthebibliography\thebibliography
\renewcommand\thebibliography[1]{
  \OLDthebibliography{#1}
  \setlength{\parskip}{0pt}
  \setlength{\itemsep}{0pt plus 0.3ex}
}

\begin{document}\sloppy

\def\x{{\mathbf x}}
\def\L{{\cal L}}

\title{Text-Guided Mask-Free Local Image Retouching}
%
\name{Zerun Liu$^1$, Fan Zhang$^1$, Jingxuan He$^2$, Jin Wang$^3$, Zhangye Wang$^2$, Lechao Cheng$^3$\sthanks{Corresponding author.}}
\address{Zhejiang University of Technology$^1$,\\ Zhejiang University$^2$, \\Zhejiang Lab$^3$}

\maketitle

\begin{abstract}
In the realm of multi-modality, text-guided image retouching techniques emerged with the advent of deep learning. Most currently available text-guided methods, however, rely on object-level supervision to confine the region of interest that may be updated. This not only makes it more challenging to develop these algorithms but also limits how widely deep learning can be used for image retouching. In this paper, we offer a text-guided mask-free image retouching approach that yields consistent results to address this concern. Specifically, we propose a two-stage mask-free training paradigm tailored for text-guided image retouching tasks. In the first stage, an unified mask is proposed according to the query description, and then several candidate images are generated with the provided mask and the conditional description based on diffusion model. Extensive experiments have shown that our method can produce high-quality images based on spoken language.

\end{abstract}
\begin{keywords}
text guided, mask free, image retouching
\end{keywords}
\section{Introduction}

With the advent of the internet and the rise in popularity of numerous photography tools, it is now easier than ever before to obtain enormous photos. In recent years, image retouching has attracted a significant deal of attention due to the need of modifying photographs to accommodate a variety of scenarios. Creating satisfying photographs using conventional techniques, however, often necessitates a substantial amount of talent and labor. Therefore, it would be convenient enough if we could modify photographs with language-specific descriptions. In this work, we focus on retouching images with language guidance.
\begin{figure}[!htb]
\centering
\includegraphics[width=\linewidth]{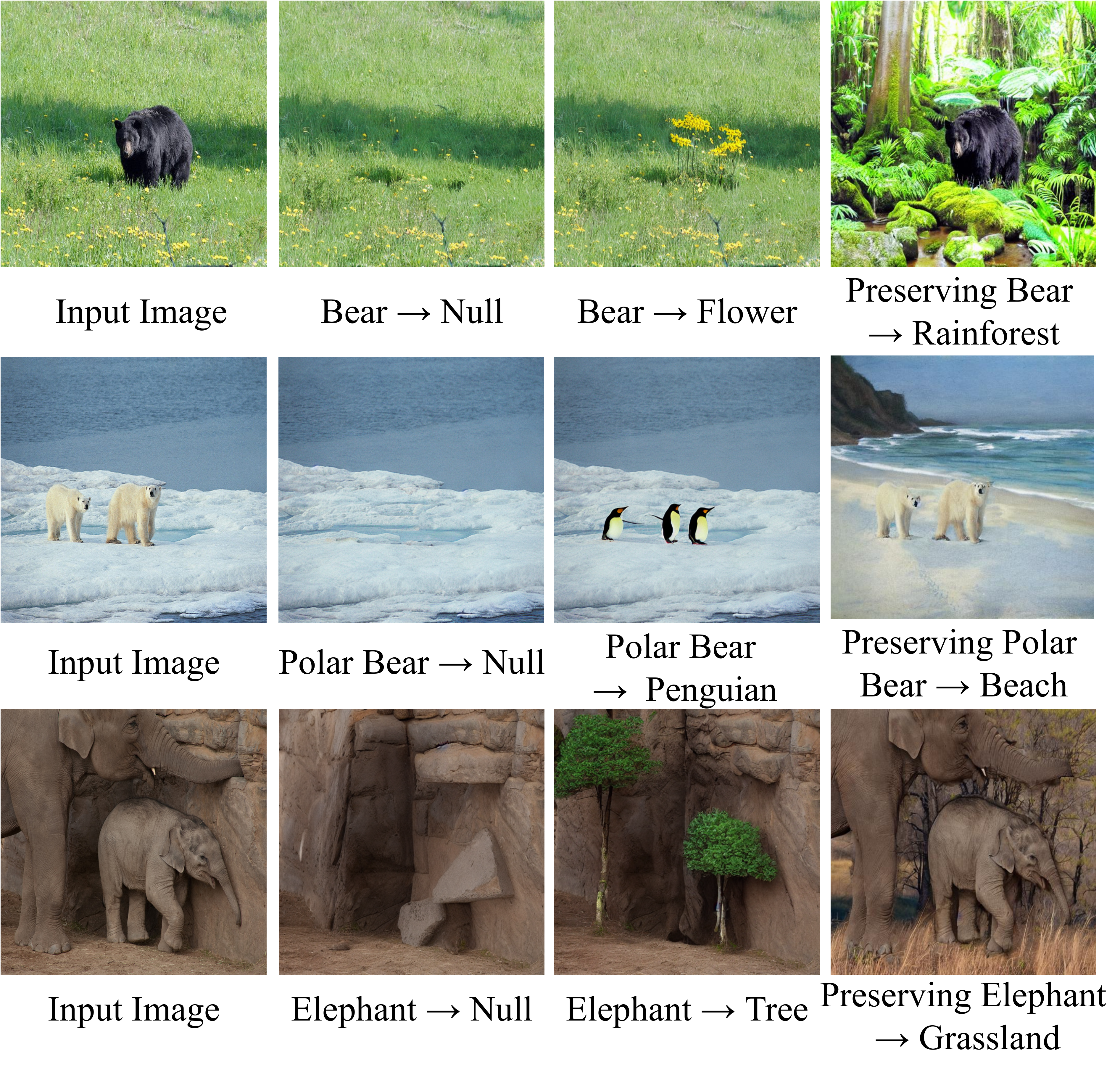}
\caption{Applications of our method: from left to right are object removal, object replacement, and background replacement.}
\vspace{-2mm}
\label{fig:teaser}

\end{figure}


Approaches for text-guided image retouching tasks can be easily classified into two categories: mask-based and mask-free methods, depending on whether the mask is used. Paint By Word~\cite{bau2021paint} is the first zero-shot solution to solve local image retouching. Following this, several diffusion-based models have manifested themselves with increasing attention, such as Blended Diffusion~\cite{avrahami2022blended}, Glide~\cite{nichol2021glide}, and Blended Latent Diffusion~\cite{avrahami2022blendedlatent}. However, it can be challenging to create an appropriate mask for each image that requires manipulation, particularly when the shape of the mask is complex.
Therefore, researchers attempt to pay more attention to mask-free techniques, which are more practical. Mask-free methods can address local image retouching without mask guidance, like VQGAN-CLIP~\cite{crowson2022vqgan}, CLIP-Guided Diffusion~\cite{ClipGuidedDiffusion}, DiffusionCLIP~\cite{kim2022diffusionclip}, and DiffuseIT~\cite{kwon2022diffusion}. However, these algorithms often suffer from the drawbacks of skewed object recognition and incomplete preservation of the original backdrop.

To cope with the above issues, we propose a two-stage framework named text-guided mask-free local image retouching, as shown in Fig~\ref{fig:teaser}, which converts a specific part of the image that matches the query to a target object described in the text. In the first stage, our method will propose an object mask according to the query. In the second step, given the input image, the suggested mask, and the text description, our algorithm will generate several candidate images, which will then be assessed to yield the final retouched image. Compared with past approaches, our method can perform local image editing without masking regions of interest. In addition, we also present the multi-modality quality assessment strategy for quantifying the quality of model outputs by picking the best one among candidate ones based on cross-model alignment and image quality assessment.


We summarize the contributions as follows:

\begin{itemize}
    \item We propose a two-stage mask-free training paradigm tailored for text-guided image retouching tasks. In the first stage, an object mask is proposed according to the query, and several candidate images are generated based on the proposed mask and the text description. Candidate images are then evaluated to produce the final retouched image.
    \item A location-aware refinement module is introduced in our framework, where users can select one entity from all entities by describing it in the text. Moreover, we also propose a multi-modal quality assessment module in which several candidate images can be assessed to help produce high-quality and semantically consistent retouched images.
    \item We conducted extensive experiments on ITMSCOCO and ITFlickr which yield promising
    visual results.
\end{itemize}

\section{Related Work}

\subsection{Text-guided Image Retouching}
The pioneering work~\cite{mansimov2015generating} introduced a deep generative model that can generate images from natural language descriptions. 
Later, another work~\cite{reed2016generative} handled image retouching by developing an effective GAN architecture conditioned on text embeddings. 
Recent advances are fueled by autoregressive transformer models that utilize VQ-VAE~\cite{van2017neural} to alleviate the problem of an unaffordable amount of computation for large models.
Diffusion models have been proven to be able to generate high-quality images, especially when natural language descriptions are given for diversity. Recent works~\cite{meng2021sdedit, nichol2021glide} explored diffusion models for text-guided image retouching and achieved promising results with efficient computations.
However, these approaches mainly specialize in global image retouching, ignoring the manipulation of a user-defined region of an image, which is a common case in our daily life.

Region-based text-guided image retouching focuses on modifying a specific region of an image while preserving the rest of it.
Paint By Word~\cite{bau2021paint} generates a realistic painting of the synthesized image based on user-provided masks and words.
Furthermore, Blended Diffusion~\cite{avrahami2022blended} was proposed for region-based editing of generic natural images instead of synthesized ones, and language descriptions are not restricted to a specific domain. 
However, the methods mentioned above require masks to guide where to edit for region-based image retouching. To tackle this problem, our framework utilizes a segmentation model to generate masks based on queries that are more convenient for users to provide.

\subsection{Diffusion Models}

Diffusion probabilistic models \cite{sohl2015deep} have recently been shown to produce high-quality images while offering sufficient image generation diversity.
Ho et al. \cite{ho2020denoising} firstly proposed to synthesize images based on diffusion models.
To take more control of the generation process, SDEdit \cite{meng2021sdedit} was presented that keeps low-frequency information by terminating the diffusion process early before it turns into a Gaussian distribution.
To further improve the quality of generated images, Dhariwal et al.~\cite{dhariwal2021diffusion} presented a trade-off between quality and diversity by adding attention pooling as a classifier. It beats some famous GAN-based image-generation methods and shows the powerful generative ability of diffusion models.
At the same time, multi-modal vision and language models trained on several large-scale datasets, such as CILP, promote the development of diffusion models by adding auxiliary information to the denoising process.
For example, Blended Diffusion~\cite{avrahami2022blended} utilizes CLIP to modify parts of the image based on language and mask guidance.
Different from Blended Diffusion, DiffusionCLIP modifies the diffusion model to match up CLIP by introducing a new loss function, which improves the diversity of the generated effect and relieves the performance of the model collapse.
Compared with classifier guidance and CLIP guidance, classifier-free guidance \cite{ho2022classifier} can acquire more accurate and better results because it can indicate an unconditional gradient estimation model and a conditional gradient estimation model at the same time in the same model. Benefiting from this, several huge diffusion models like Glide \cite{nichol2021glide}, DALLE2 \cite{ramesh2022hierarchical}, Imagen \cite{saharia2022photorealistic} are proposed and make amazing results.
\begin{figure*}[!htb]
\begin{center}
\includegraphics[width=\linewidth]{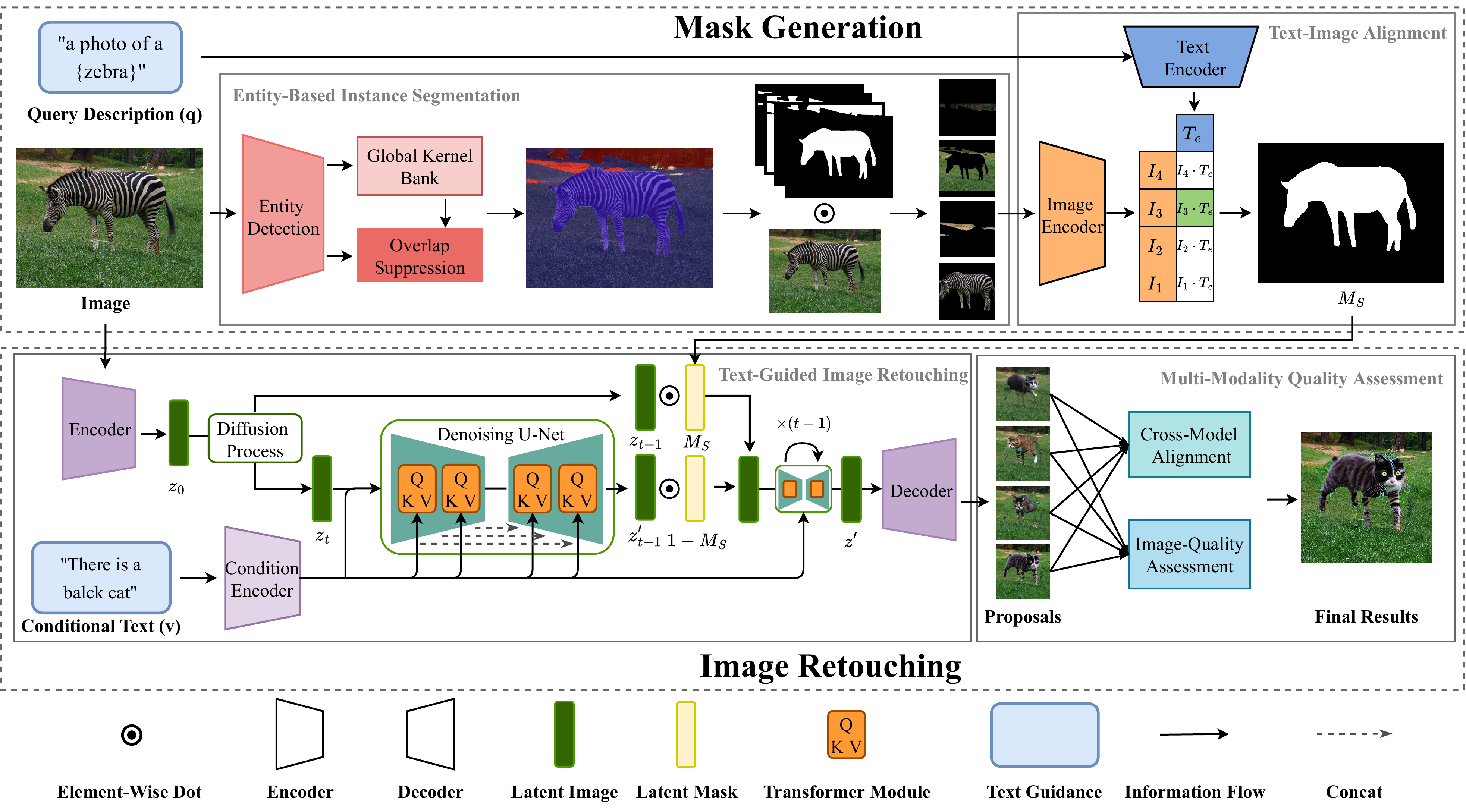}
\end{center}
\caption{The overview of our framework. It comprises two parts: image retouching and mask generation. Based on the query, mask generation tries to create the mask based on the query description. Image retouching focuses on altering the image based on the provided conditional text and picking the best of several proposals via the multi-modality quality assessment module.}
\label{fig:Framework}
\end{figure*}

\section{Method}

\subsection{Overview}
Let $I \in \mathbb{R}^ {H\times W\times 3}$ be the input image, $\mathbf{q}$ be the query description and $\mathbf{v}$ be the conditional text, respectively. The goal of the proposed framework attempts to retouch $I$ at the local region of interest based on $\mathbf{q}$ and $\mathbf{v}$ to produce a high-quality natural image. 

To achieve this, we draw inspiration from entity segmentation~\cite{qi2021open} to segment all visual entities $I_i, i\in \left \{0, ..., n-1 \right \}$ in an image based on mask potentials $M_i$ without considering semantic category labels. Each visual entity $I_i$ is generated as follows:
\begin{equation}
  I_{i} = I \odot M_i
\label{equ:getEntity}
\end{equation} 
Where $\odot $ denotes an element-wise multiplication operation. 

Then, we adopt pre-trained CLIP models to predict the correct pairings of a batch of (image, text) examples. Assuming that $\mathbf{E}_T$ and $\mathbf{E}_I$ are text encoder and image encoder, respectively. For each entry in multi-modal embedding space spanning by visual entities and query descriptions, we compute the correlation $\mathcal{C}_i$ as:
\begin{equation}
    \mathcal{C}_i = \mathbf{E}_I(I_i) \cdot \mathbf{E}_T(\mathbf{q})
\end{equation}\label{eq:corr}
Notably that we consider all query words as one instance to form single embedding $T_e=\mathbf{E}_T(\mathbf{q})$, as illustrated in Fig~\ref{fig:Framework}. To determine the regions that ought to be edited, the mask $\mathbf{M}_S$ with high correlation confidence is selected. Formally, $\mathbf{M}_S = \bigcup_{i}M_i , ~st.~ \mathcal{C}_i \geq \tau$. In practice, fixing threshold $\tau$ does not work well in real sceneries, and we introduce an adaptive thresholding strategy inspired by~\cite{su2022re}. More details can be referred at Sec~\ref{sec:iea}.

Once we obtain the accurate regions of interest, the input image $I$, and the conditional text $\mathbf{v}$ as well as detected areas $\mathbf{M}_S$ are fed into a diffusion model to generate retouched proposals $\mathbf{P}_k, k\in \{0, ..., m-1\}$. $m$ is the number of proposals. We finally attach a multi-modality quality assessment module to produce high-quality and semantic-aware results based on proposed Cross-Model Alignment (CMA) and Image-Quality Assessment (IQA) strategies. The subsequent sections will detail the above components.

\subsection{Mask Generation}
Unlike existing popular local text-guided image retouching techniques, our model will construct the mask based on the query to determine where to modify. In particular, it can generate accurate masks in relatively simple scenarios, however, it fails when it comes to more complex situations. In this section, we attempt to generate more precise masks based on the query description. We will outline them in detail below. 

\subsubsection{Entity-Based Instance Segmentation}
The entity-based instance segmentation module is built upon ES~\cite{qi2021open}, a novel open-world entity segmentation task to investigate the feasibility of convolutional center-based representation to segment things and stuff in a unified manner. The introduced global kernel bank and overlap suppression aim to exploit the requirements of ES in order to improve the segmentation quality of predicted entity masks. As a result, the mask potentials $M_i$ and visual entities $I_i$ are obtained for subsequent processing.
\subsubsection{Text-Image Alignment}\label{sec:iea}
We leverage the CLIP~\cite{radford2021learning} strategy to model cross-modality relations for query description $\mathbf{q}$ and visual entities $I_i$. As stated in eq~\ref{eq:corr}, $\tau$ plays an important role in obtaining precise regions. Thus, we adopt an adaptive thresholding strategy based on sampling over cumulative correlation confidence. Specifically, we first sort the correlation values in $\mathcal{C}_i, i\in \{0, ..., n-1\}$ from high to low and calculate the cumulative confidence. Then the adaptive threshold $\tau$ is obtained based on the inverse transform sampling over the maximal contribution step. 

\subsubsection{Location-Aware Refinement}
The above text-image alignment procedure can work well in most cases, but it may fail when descriptions contain location constraints. To tackle this issue, we propose a location-aware refinement module that amends the mask region as expected. We simply split the image space into nine regions evenly and bag each detected visual entity into the corresponding location. Recall that we separate the regions and background based on the correlation confidence $\mathcal{C}$, and we proceed to perform a location-aware refinement operation by restricting the orientation. 

\subsection{Text-Guided Image Retouching}
After obtaining regions $\mathbf{M}_S$ correlated with query description $\mathbf{q}$, we further retouch the image $I$ with conditional text $\mathbf{v}$ with provided mask $\mathbf{M}_S$. The input image $I$ and the conditional text $\mathbf{v}$ will be compressed into the same latent space, respectively. Formally, denoting $z_0$ as the initial latent vector, we utilize the denoising network $\epsilon_{\theta}(z_t, t, \mathbf{v})$ to reconstruct $z$ from the Gaussian distribution $\mathcal{N}(0,I)$. Thus, we can simplify the corresponding objective as:
\begin{equation}
  \mathcal{L} := \mathbb{E}_{z,\epsilon \sim \mathcal{N}(0,I),t} \left [ \left \| \epsilon - \epsilon_{\theta}(z_{t}, t, \mathbf{v}) \right \| \right ]
\label{equ:latentCorrespondingObjectivesves}
\end{equation}
For timestep $t$, the denoising result $z^{'}_{t}$ is further updated with mask $\mathbf{M}_S$ as:
\begin{equation}
  z^{'}_{t}=z_{t}\odot (1 - \mathbf{M}_S) + z^{'}_{t}\odot \mathbf{M}_S
\label{equ:combineTwoPorcess}
\end{equation}
Eq~\ref{equ:combineTwoPorcess} indicates that we keep the original background while learning text-guided visual entities based on the conditional text $\mathbf{v}$. The last $z^{'}_{t}$ is put forward into a decoder network to generate plausible proposals $\mathbf{P}_k, k\in \{0, ..., m-1\}$.

\subsection{Multi-Modality Quality Assessment}
Randomization is an essential factor of the denoising procedure, which assures the variety of the diffusion model, yet it may also lead to undesirable outcomes. To address this issue, we present a multi-modality quality assessment module for selecting the best result from several candidates. In this module, we examine the quality of the output from two perspectives: cross-model assessment and image quality assessment.
\subsubsection{Cross-Model Alignment}
In practice, we employ CLIP to measure the consistency between proposals $\mathbf{P}_k$ and the conditional text $\mathbf{v}$ as described in section~\ref{sec:iea}. The cross-model confidence $\mathcal{C}'_k$ for proposal $\mathbf{P}_k$ is then defined as:
\begin{equation}
    \mathcal{C}'_k = \mathbf{E}_I(\mathbf{P}_k) \cdot \mathbf{E}_T(\mathbf{v})
\end{equation}
Noted that $\mathcal{C}'_k$ is normalized to $[0, 1]$. Recall that $\mathbf{E}_T$ and $\mathbf{E}_I$ are text encoder and image encoder before.
\subsubsection{Image Quality Assessment}
In addition to cross-modality alignment, we shall also pay attention to the visual quality of the generated result. Therefore, we introduce an image quality assessment module to enable high-quality output images. Formally, we define the image quality confidence score $\mathcal{S}_k$ as follows:
\begin{equation}
   \mathcal{S}_k = \frac{1}{H*W}\sum_i\sum_j \| I(i, j) - \mathbf{P}_k(i, j)\|
\end{equation}

Finally, we choose the best suitable proposal $\mathbf{P}_{k'}$ that fulfills:
\begin{equation}
    k' := \argmax_k ( \mathcal{C}'_k - \alpha\cdot\mathcal{S}_k )
\end{equation}
Where $\alpha$ is the hyperparameter, and we empirically set $\alpha=5.0$. Experiments indicate they can increase the overall quality of image retouching results with semantic consistency.

\section{Experiment}
\subsection{Experimental Settings}
We follow~\cite{luo2018discriminability} to generate ITMSCOCO and ITFlickr by selecting 5000 text-image pairs from MSCOCO~\cite{lin2014microsoft} and Fliker30K~\cite{young2014image}, respectively. Furthermore, SSIM, PSNR, FID, MSE, and LPIPS are utilized as metrics to quantify the quality of retouching results.
\subsection{Compared to Existing Methods}

\begin{figure*}[htb]
\centering
\resizebox{0.95\linewidth}{!}{%
\includegraphics[clip, trim={3cm, 1cm, 4cm, 0}, width=0.3\textwidth]{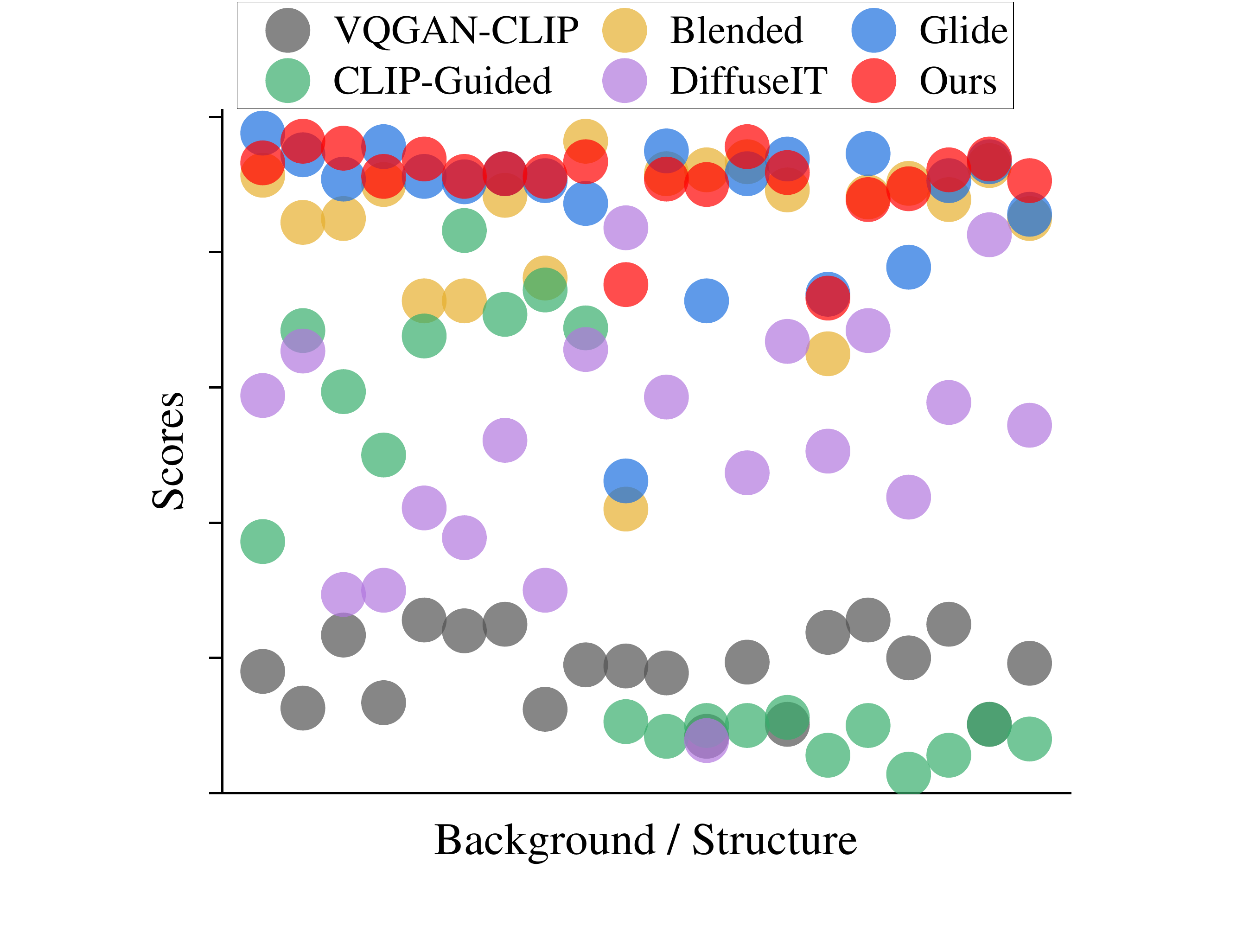}
\includegraphics[clip, trim={3cm, 1cm, 4cm, 0}, width=0.3\textwidth]{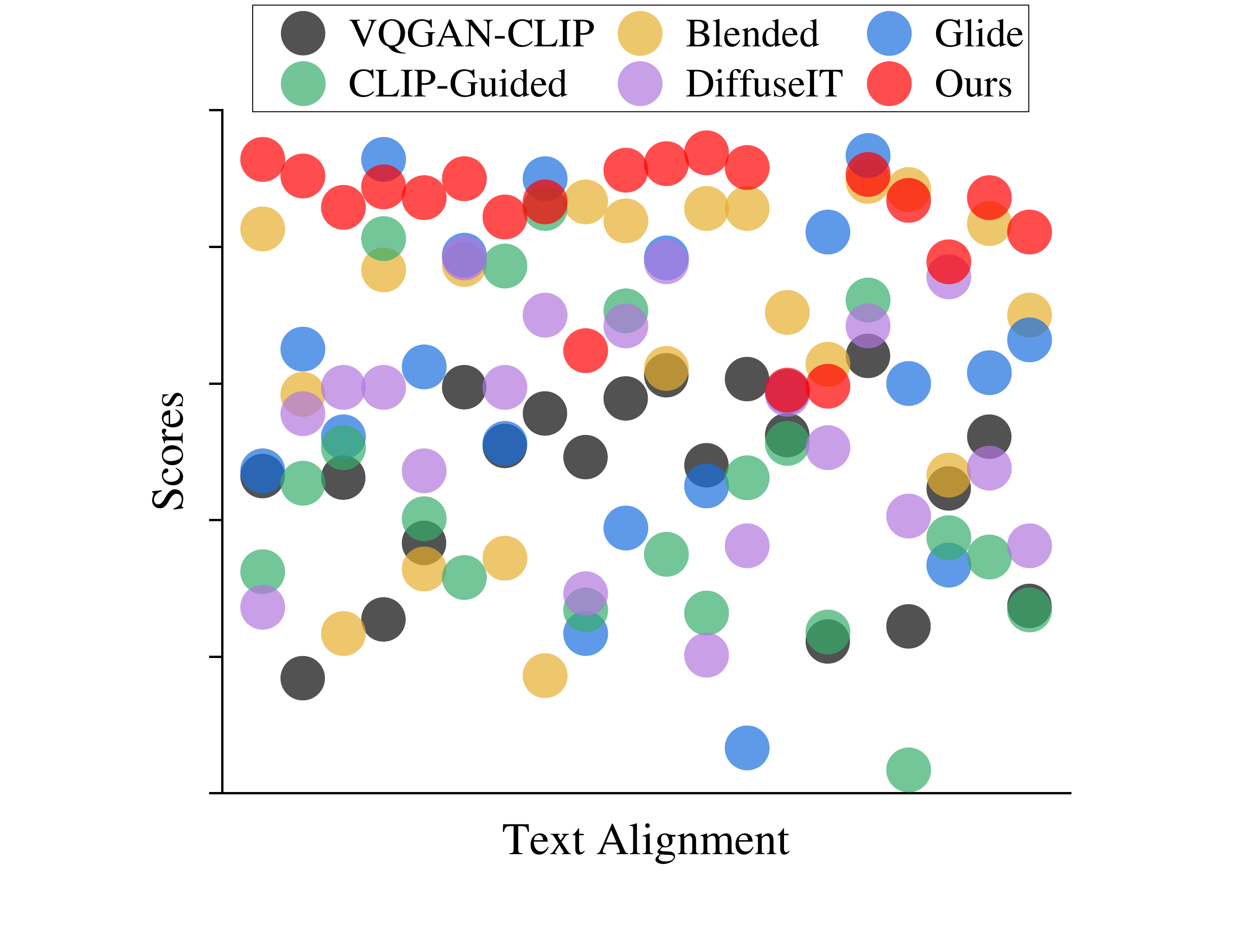}
\includegraphics[clip, trim={3cm, 1cm, 4cm, 0}, width=0.3\textwidth]{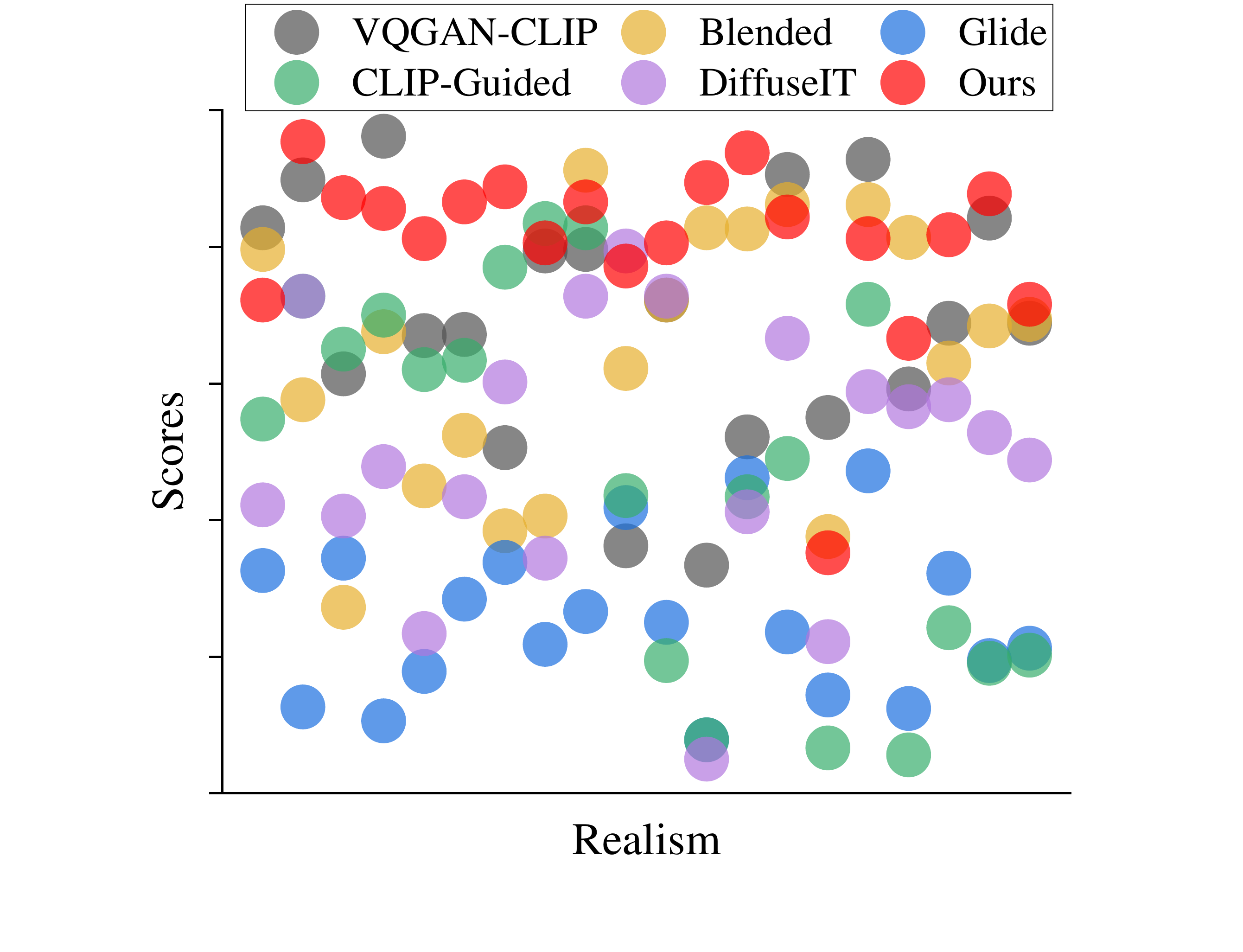}
}
\vspace{-0.5cm}
\caption{After each image was assessed by 20 participants, we computed the mean scores of \textbf{Background / Structure}, \textbf{Text Alignment} and \textbf{Realism}) for 20 generated images. }
\label{fig:user_study}  
\vspace{-0.6cm}
\end{figure*}

To evaluate the performance of our method, we compare our results against the following state-of-the-art baseline models: Glide, Blended Latent Diffusion, VQGAN-CLIP, CLIP-Guided Diffusion, and DiffuseIT. Experiments show that our model outperforms the above methods qualitatively and quantitatively. 

\subsubsection{Qualitative Comparison}
\begin{figure*}[!htb]
\centering
\includegraphics[clip, trim=0 8cm 0 0cm, width=\linewidth]{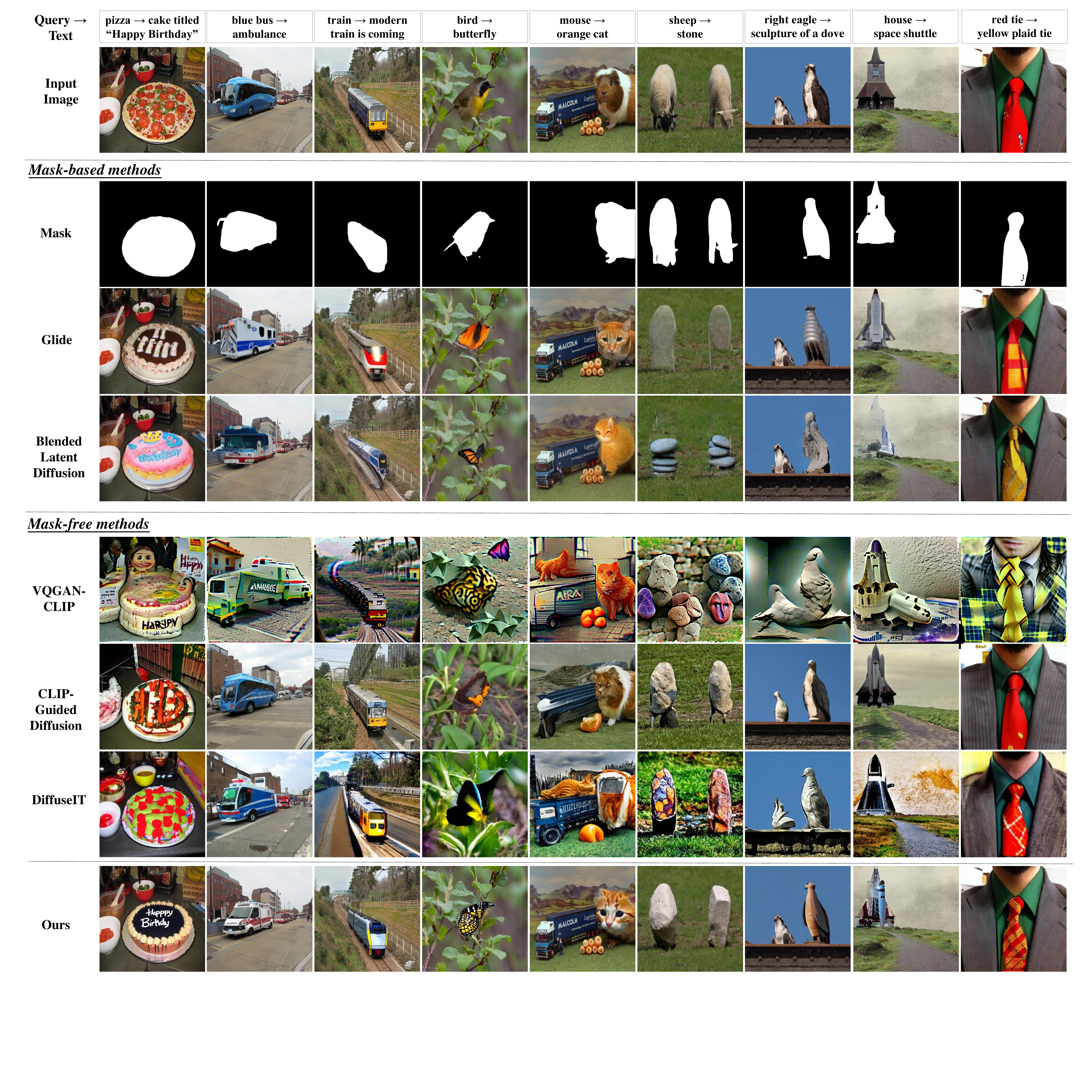}
\caption{Qualitative comparison of text-guided image retouching on MSCOCO dataset. We choose two mask-based methods and three mask-free methods. Our approach generates realistic samples based on the conditional text with better perceptual quality than the baselines.}
\label{fig:ImageAblation}
\end{figure*}

As shown in Fig~\ref{fig:ImageAblation}, we compare our results with two mask-based methods and three mask-free methods. The masks for the mask-based methods are generated by our method, which shows our superiority that we needn't provide the mask for the regions of interest. As for mask-free methods, the main drawback of them is that they couldn't preserve the whole background well, which may lead to the failure of retouching, besides, these methods may fail when handling complex situations like Column 7 of Fig~\ref{fig:ImageAblation}. As a result, our method brings the best of both worlds and obtains persuasive results.
\subsubsection{Quantitative Comparison}
In the realm of text-guided image retouching, it's difficult to measure the performance of models by metrics quantitatively. Therefore, we introduce a user study, as depicted in Table~\ref{tab:userstudy} and Fig.~\ref{fig:user_study}. The study participants were instructed to evaluate each result based on (1) the preservation of the background and structure relative to the source image, (2) alignment with the corresponding text, and (3) the realism of the output. A total of 20 participants took part in the study and evaluated 20 randomly selected examples for each method. The examples were presented in random order. For each example, the participants rated the quality on a scale of 0 to 5, with higher ratings indicating better performance.

\vspace{-5mm}
\begin{table}[!htb]
\begin{center}
\caption{The quantitative statistical rating of generated results. $\uparrow$ means that the higher the value, the better. \textcolor{red}{Red} indicates the best scores, while \textcolor{blue}{blue} indicates the second-best results.} \label{tab:userstudy}
\resizebox{0.99\linewidth}{!}{%
\begin{tabular}{c|c|c|c}
  \hline
  \multirow{2}{*}{Methods} & Background / & \multirow{2}{*}{Text Alignment ($\uparrow$)} & \multirow{2}{*}{Realism ($\uparrow$)} \\
   & Structure ($\uparrow$) & & \\
    \hline
Glide~\cite{nichol2021glide} & \textcolor{blue}{4.36±0.59} & 2.97±1.16 & \textcolor{blue}{3.47±0.9} \\   \hline
Blended~\cite{avrahami2022blendedlatent} & 4.19±0.64 & \textcolor{blue}{3.29±1.16} & 3.23±0.95 \\   \hline
VQGAN~\cite{crowson2022vqgan} & 0.93±0.28 & 2.25±0.73 & 1.27±0.58 \\   \hline
Clip-guided~\cite{ClipGuidedDiffusion} & 1.68±1.5 & 2.25±1.11 & 2.35±1.35 \\   \hline
DiffuseIT~\cite{kwon2022diffusion} & 2.58±0.94 & 2.66±0.87 & 2.5±0.96\\   \hline
\textbf{Ours} & \textcolor{red}{4.52±0.3} & \textcolor{red}{4.21±0.53} & \textcolor{red}{4.03±0.65} \\ 
  \hline
\end{tabular}
}
\end{center}
\vspace{-0.8cm}
\end{table}


\subsection{Ablation Studies}
In this section, we conduct ablation studies to verify the effectiveness of the mask generation and image retouching modules.
\subsubsection{Mask generation}
\begin{figure}[!htb]
\centering
\includegraphics[width=\linewidth]{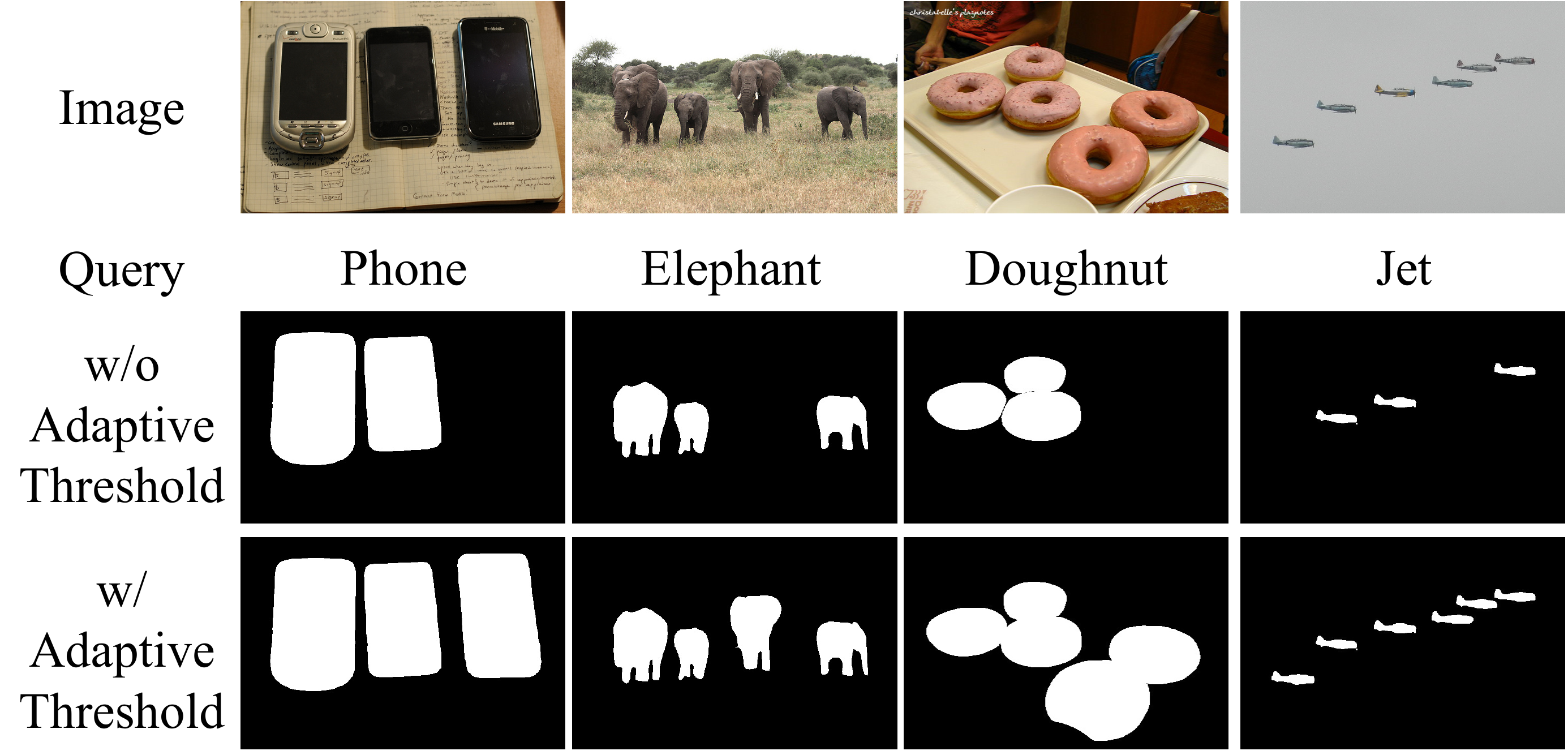}
\caption{Adaptive threshold. We empirically set the threshold to $\tau$ to 0.2 for the baseline.}
\label{fig:DynamicThreshold}
\end{figure}

\noindent \textbf{Adaptive Threshold} As illustrated in Fig~\ref{fig:DynamicThreshold}, the empirically rough threshold often cause the loss of visual entities when complicated scenarios are encountered. While our adaptive threshold exhibits much more characteristics that capture the regions of interest well.

\begin{figure}[!htb]
\centering
\includegraphics[width=\linewidth]{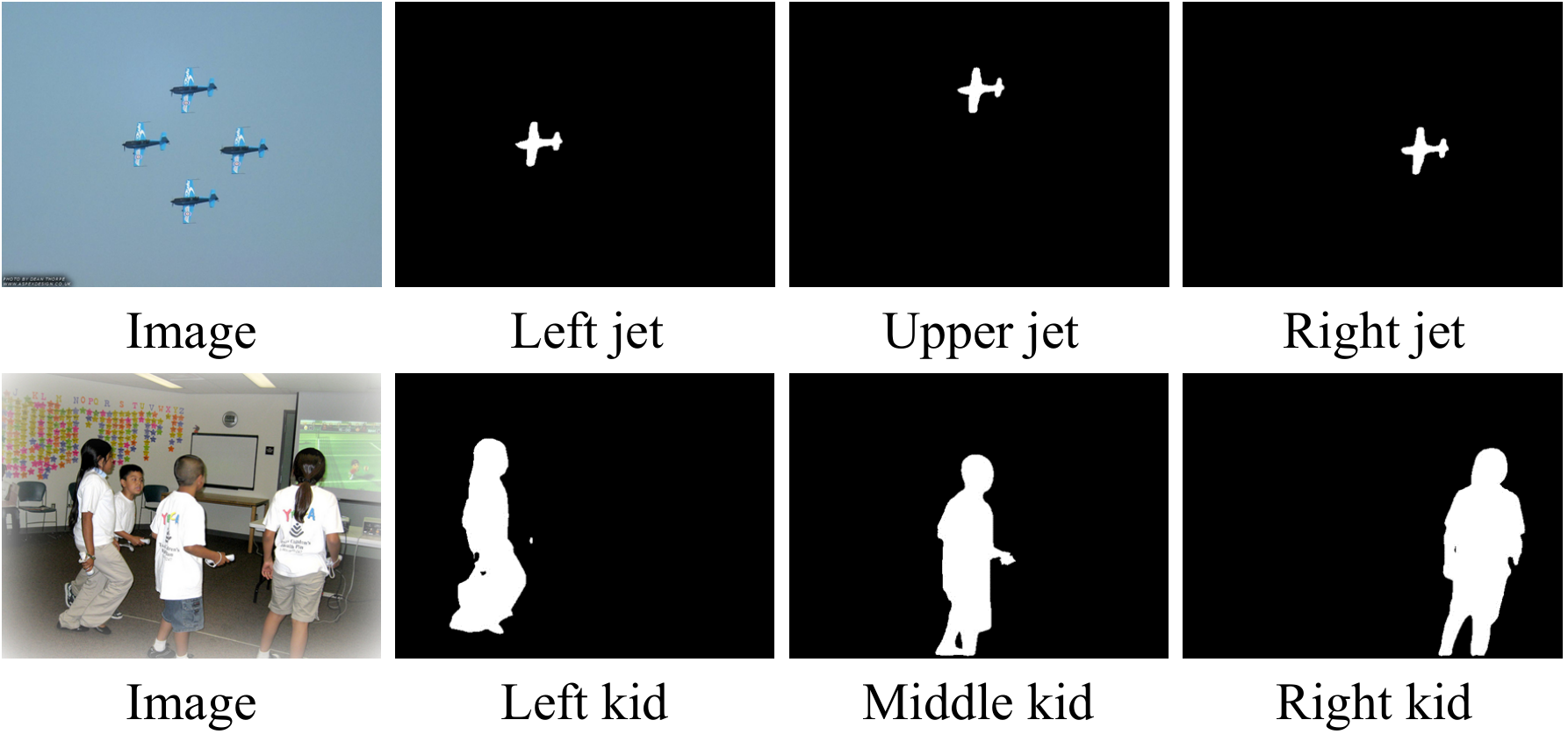}
\caption{Visual entities with orientation.}
\vspace{-2mm}
\label{fig:DirectionMaskGeneration}
\end{figure}

\noindent \textbf{Location-Aware Refinement} We demonstrate the effectiveness of the proposed location-aware refinement in Fig~\ref{fig:DirectionMaskGeneration} and most common positional restraints are well retained in real cases.

%
%
\subsubsection{Multi-Modality Quality Assessment}
\noindent\textbf{CMA \textit{vs.} IQA}.

\begin{figure}[!htb]
\centering
\includegraphics[width=\linewidth]{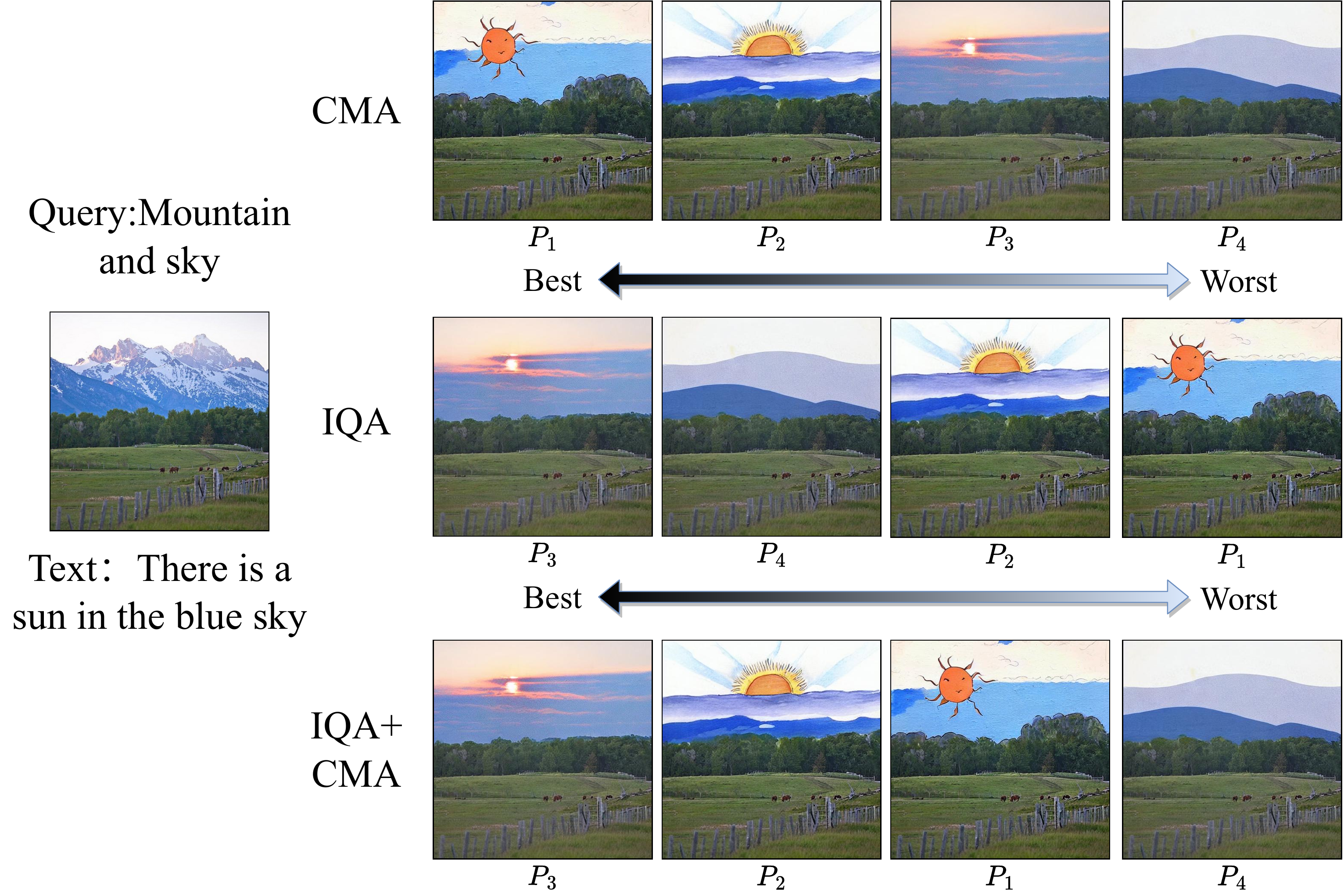}
\caption{The effectiveness of CMA and IQA. $\mathbf{P}_1, \mathbf{P}_2, \mathbf{P}_3, \mathbf{P}_4$ are ranked only based on CMA strategy, while $\mathbf{P}_3, \mathbf{P}_4, \mathbf{P}_2, \mathbf{P}_1$ are ranked by the IQA module. After combining the CMA and IQA module, the final result changed to $\mathbf{P}_3, \mathbf{P}_2, \mathbf{P}_1, \mathbf{P}_4$.}
\vspace{-2mm}
\label{fig:IQAvsCMA2}

\end{figure}

As can be seen in Tab~\ref{tab:twoModules}, we evaluate the CMA and IQA modules with the mean values of all validation images. The quantitative results just reveal that the IQA component plays an essential role in the production of high-quality images, while the qualitative visual results in Fig~\ref{fig:IQAvsCMA2} show us the significant involvement of CMA to yield semantic-aware targets.
\begin{table}[!htb]
\centering
\caption{CMA \textit{vs.} IQA. We carry out comparative experiments on the ITMSCOCO datasets. The max timestep is 200, and the proposal number $m$ equals 4.}  
\label{tab:twoModules}  
\vspace{0.5mm}
\resizebox{\columnwidth}{!}{%
\begin{tabular}{c|c|c|c|c|c|c}
\hline
CMA & IQA & SSIM($\uparrow$) & PSNR($\uparrow$) & FID($\downarrow$) & MSE($\downarrow$) & LPIPS($\downarrow$)\\ \hline
 $\times$   &  $\times$  &   0.7042   &  17.7389   &  16.7268  & 0.0218  &  0.2673  \\ \cline{1-7} 
 $\surd$   &  $\times$  &   0.7121   &  17.7823   &  14.5790  & 0.0191  &  0.2608  \\ \cline{1-7} 
 $\times$  &  $\surd$   &   0.7218   &  18.6803   &  12.7913  & 0.0160  &  0.2499  \\ \cline{1-7} 
 $\surd$   &  $\surd$   &   0.7125   &  17.8141   &  14.5512  & 0.0190  &  0.2605  \\ \cline{1-7}
\end{tabular}%
}
\end{table}



\section{Conclusion}
This study offers a two-stage framework for text-guided mask-free local image retouching. This framework transforms a specified region of the image that fits the query into a text-described target object. Extensive experiments demonstrate the superiority of the proposed approach against existing methods.

\section*{Acknowledgments}
This work is partially supported by the National Natural Science Foundation of China (Grant No. 62106235), by the Exploratory Research Project of Zhejiang Lab(2022PG0AN01), by the Zhejiang Provincial Natural Science Foundation of China (LQ21F020003).

\bibliographystyle{IEEEbib}
\bibliography{reference}

\end{document}